\documentclass{article}
\usepackage[dvipsnames]{xcolor} 
\usepackage{microtype}
\usepackage{graphicx} 
\usepackage{subfigure}
\usepackage{booktabs} 
\usepackage{fancyhdr} 
\usepackage{float}
\usepackage{multirow} 

\usepackage{hyperref}
\hypersetup{
    colorlinks=true,
    linkcolor=violet,
    citecolor=YellowOrange,
    urlcolor=Aquamarine
}

\usepackage[accepted]{icml2021}

\usepackage{amsmath}
\usepackage{amsfonts}
\usepackage{algpseudocodex}
\usepackage{caption}       
\usepackage[framemethod=TikZ]{mdframed} 
\usepackage{authblk}       

\usepackage[T1]{fontenc}
\usepackage{multicol}
\usepackage{enumitem}

\usepackage[english]{babel} 

\usepackage{cleveref}
\icmltitlerunning{Adaptive Graph of Thoughts}

\begin{document}

\pagestyle{fancy} 
\fancyhf{} 
\cfoot{\vspace{1cm} \thepage} 

\twocolumn[
\icmltitle{Adaptive Graph of Thoughts: Test-Time Adaptive Reasoning Unifying Chain, Tree, and Graph Structures}

\begin{icmlauthorlist}
\icmlauthor{\href{https://www.linkedin.com/in/tpmath/}{Tushar Pandey}}{agn}
\icmlauthor{\href{https://www.linkedin.com/in/ara-ghukasyan/}{Ara Ghukasyan}}{agn}
\icmlauthor{\href{https://www.linkedin.com/in/oktay-goktas-8b2881167/}{Oktay Goktas}}{agn}
\icmlauthor{\href{https://www.linkedin.com/in/santoshkumarradha/}{Santosh Kumar Radha}}{agn}
\end{icmlauthorlist}

\icmlaffiliation{agn}{Agnostiq Inc., 325 Front St W, Toronto, ON M5V 2Y1}

\icmlcorrespondingauthor{Santosh Kumar Radha}{contact@agnostiq.ai}

\printAffiliationsAndNotice{}

\begin{center}
    Agnostiq Inc., 325 Front St W, Toronto, ON M5V 2Y1
\end{center}

\vskip 0.2in

\printAffiliationsAndNotice{}

\begin{abstract}
Large Language Models (LLMs) have demonstrated impressive reasoning capabilities, yet their performance is highly dependent on the prompting strategy and model scale. While reinforcement learning and fine-tuning have been deployed to boost reasoning, these approaches incur substantial computational and data overhead. In this work, we introduce Adaptive Graph of Thoughts (AGoT), a dynamic, graph-based inference framework that enhances LLM reasoning solely at test time. Rather than relying on fixed-step methods like Chain of Thought (CoT) or Tree of Thoughts (ToT), AGoT recursively decomposes complex queries into structured subproblems, forming an \textit{dynamic} directed acyclic graph (DAG) of interdependent reasoning steps. By selectively expanding only those subproblems that require further analysis, AGoT unifies the strengths of chain, tree, and graph paradigms into a cohesive framework that allocates computation where it is most needed. We validate our approach on diverse benchmarks spanning multi-hop retrieval, scientific reasoning, and mathematical problem-solving, achieving up to 46.2\% improvement on scientific reasoning tasks (GPQA) - comparable to gains achieved through computationally intensive reinforcement learning approaches and outperforming state-of-the-art iterative approaches. These results suggest that dynamic decomposition and structured recursion offer a scalable, cost-effective alternative to post-training modifications, paving the way for more robust, general-purpose reasoning in LLMs.
\end{abstract}

]


\begin{figure}[!ht]
    \centering
    \includegraphics[width=0.7\linewidth]{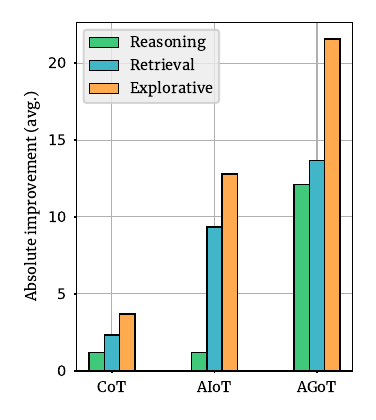}
    \caption{Performance comparison of reasoning frameworks (Chain of Thought, Autonomous Iteration of Thought, and Adaptive Graph of Thoughts) against input-output baseline using \texttt{gpt-4o-mini}. Bars represent absolute improvement in percentage points across reasoning, retrieval, and explorative task categories. AGoT demonstrates consistent performance gains across all categories, with highest improvements in explorative tasks. (see \autoref{subsubsec:gpqa}).}
    \label{fig:performance comparison 1}
\end{figure}

\section{Introduction}
\begin{figure*}[!htb]
    \centering
    \includegraphics[width=1\linewidth]{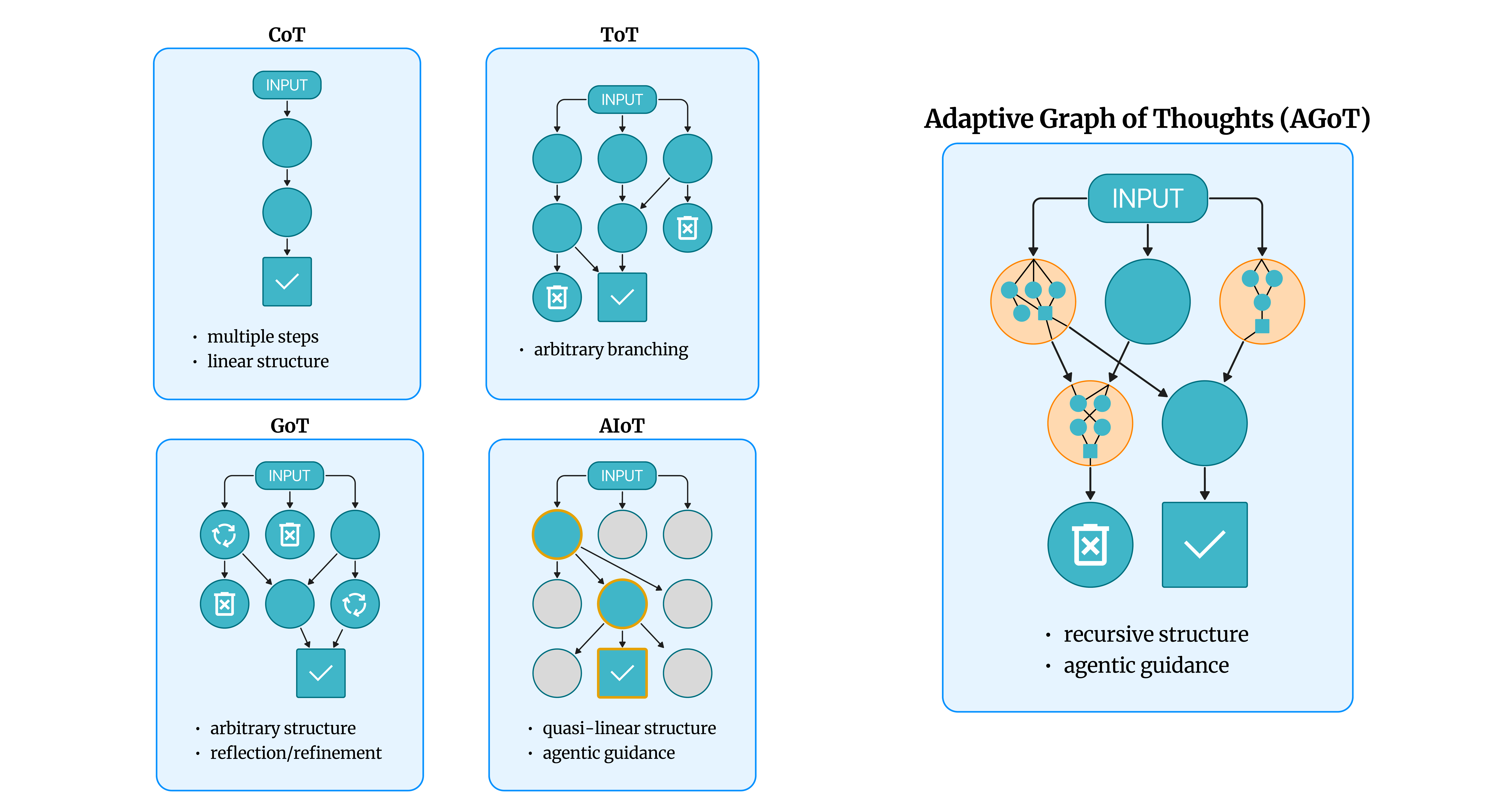}
    \caption{Architectural comparison of inference frameworks showing structural evolution from linear (CoT) to more complex reasoning patterns. Chain of Thought (CoT)~\citep{wei2022chain} employs sequential reasoning, Tree of Thoughts (ToT)~\citep{yao2024tree} introduces branching pathways, Graph of Thoughts (GoT) enables arbitrary connections with refinement, and Autonomous Iteration of Thought (AIoT)~\citep{radha2024iterationthoughtleveraginginner} implements quasi-linear structure with dynamic guidance. The proposed AGoT framework unifies these approaches through recursive graph structures (orange circles indicating nested computation) while maintaining directed acyclic connectivity. Node symbols: circles represent thought states, squares with checkmarks indicate final answers, and crossed squares denote terminated paths.}
    \label{fig:agot-hero}
\end{figure*}
Large Language Models (LLMs) have rapidly advanced the state of natural language processing, demonstrating impressive capabilities in understanding, reasoning, and problem-solving. Despite these advances, the quality of inference produced by LLMs remains highly sensitive to the prompting strategy and model scale. Traditional techniques such as reinforcement learning and fine-tuning~\citep{dubey2024llama, openai2024o1} have been successful in enhancing reasoning performance, achieving up to +46\% improvement on standard benchmarks through computationally intensive model distillation~\citep{deepseekai2025deepseekr1incentivizingreasoningcapability}; however, these methods incur substantial computational and data overhead. In this work, we propose an inference-time alternative—\emph{Adaptive Graph of Thoughts} (AGoT)\footnote{An installable implementation of the AGoT framework and result data can be found at~\citep{multi_agent_llm_2024} Github.}—which achieves comparable performance gains (+46.2\% on GPQA) (\cref{tab:gpt-gpqa-results}) through dynamic decomposition of complex queries into structured subproblems. By unifying the sequential nature of Chain of Thought (CoT)~\citep{wei2022chain}, the branching strategies of Tree of Thoughts (ToT)~\citep{yao2024tree}, and the generality of graph-based reasoning~\citep{besta2024graph}, AGoT offers a scalable and cost-effective framework for enhancing LLM reasoning without modifying the base model. Our approach demonstrates that careful structuring of the inference process can match the benefits of computationally intensive training methods while preserving model architecture and reducing computational overhead.

\subsection{Techniques for LLM Improvement}

Advances in training techniques, dataset selection, and neural network architectures continue to drive the rapid development of artificial intelligence (AI) and LLMs~\citep{anil2023palm, openai2023gpt4, team2023gemini, dubey2024llama, openai2024o1}. As LLM-powered AI permeates the software ecosystem, understanding user-AI or AI-AI interactions becomes increasingly important for building trust in applications. LLMs are conceptually opaque~\citep{singh2023explainingblackboxtext}, with both beneficial and detrimental subtleties in their behavior still emerging from empirical studies~\citep{ajwani2024llmgeneratedblackboxexplanationsadversarially}. Inference frameworks have thus emerged as systematic solutions for enhancing high-level LLM interactions.

Chain of Thought (CoT)~\citep{wei2022chain} was a pioneering example of systematic prompting, demonstrating that LLM reasoning can be improved when prompts include explicit instructions to follow sequential logical steps. Building upon this insight, subsequent methods have incorporated iterative query refinement~\citep{krishna2024understandingeffectsiterativeprompting, radha2024iterationthoughtleveraginginner} and branching thought patterns~\citep{wang2022self, yao2024tree, besta2024graph} and multi-agent systems \citep{radha2024composite} to elevate performance further. More recent frameworks, such as AIoT~\citep{radha2024iterationthoughtleveraginginner}, aim to boost generalization by dynamically selecting thought patterns and enabling greater flexibility in structuring the reasoning process. 

In this context, we introduce \emph{Adaptive Graph of Thoughts} (AGoT) as a natural successor. Leveraging dynamic decomposition via directed acyclic graphs (DAGs) in a manner akin to ToT~\citep{yao2024tree} and GoT~\citep{besta2024graph}, AGoT distinguishes itself through the incorporation of LLM-driven complexity checks. These checks enable the recursive evaluation of complex tasks and sub-tasks via nested graphs. \cref{fig:agot-hero} visually compares AGoT to several of its predecessors.

\subsection{Thought structures as graphs}

\begin{figure}
    \centering
    \includegraphics[width=1\linewidth]{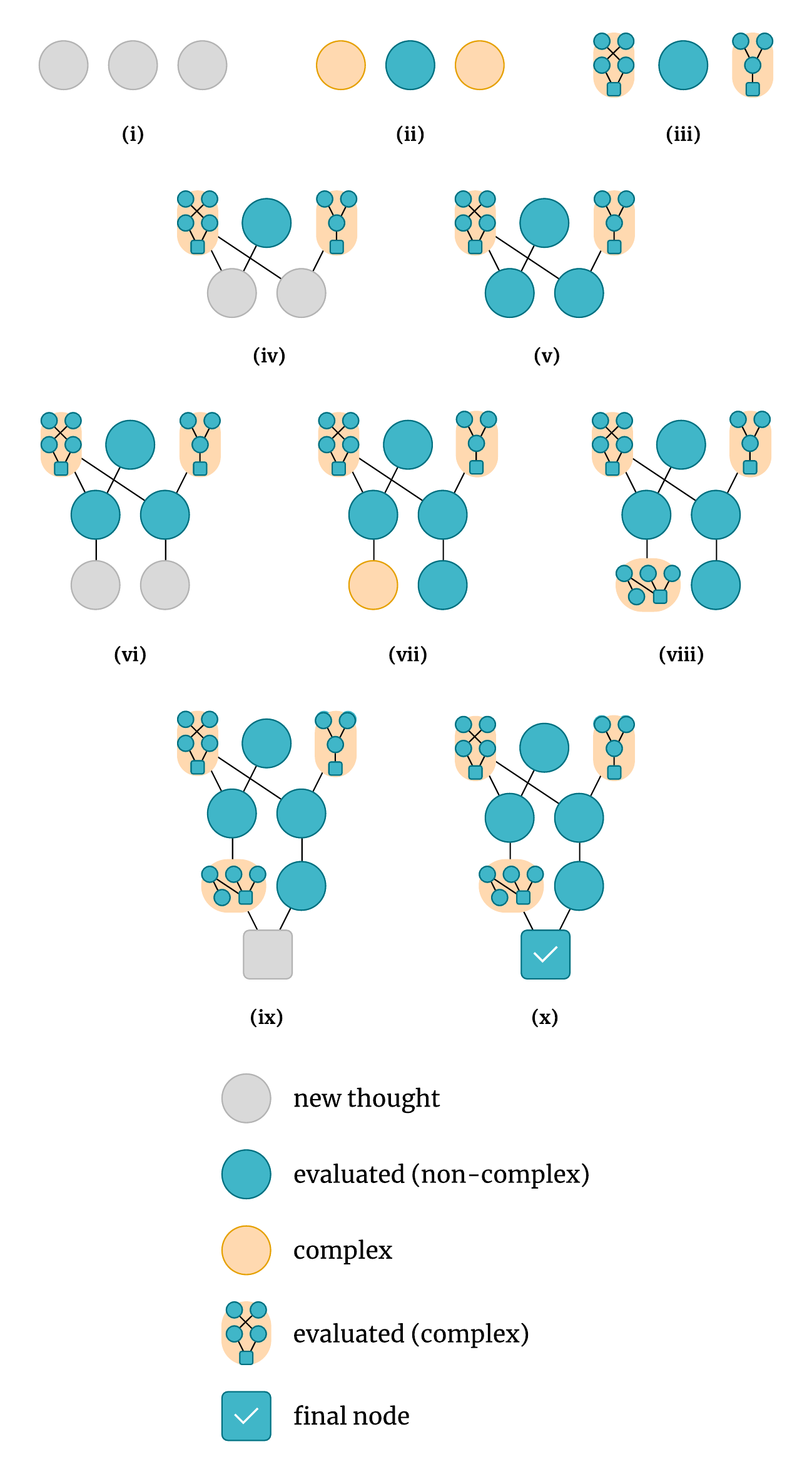}
    \caption{Schematic showing the layer-wise evolution of AGoT. Each row depicts the creation and completion of a single layer. Miniature nested graphs are superimposed on complex nodes to indicate completion. A check symbol identifies the final node in the top-level graph. All edges are directed downward across neighboring generations. Edges entering complex nodes are implicitly connected to every node in the first layer of the nested graph. Edges exiting complex nodes implicitly originate from the final node of the nested graph.}
    \label{fig:agot-steps}
\end{figure}

We understand a "thought" to be any unit of content that informs an LLM's final response. Thoughts are assumed to be accessible as combinable units of information rather than any implicit steps in an LLM's internal reasoning process. Given a set of tangible thoughts, we construct a graph where each directed edge indicates that one thought informs another. Edge traversal in this graph is then analogous to a traceable thought process, and considering only \textit{acyclic} graphs ensures all thought processes are finite. Any DAG of thoughts, whether chain- or tree-like or otherwise, contains one or more thought processes as identified by the ancestors of an arbitrary "final" node.

The evolution of the AGoT framework is governed by the gradual expansion of this graph. Starting from an empty graph, node and edge generation proceeds one topological layer at a time, with the thoughts at each layer undergoing evaluation through LLM queries that are contextualized by the current state of the graph. The complete evolution of a sample AGoT graph is shown in \autoref{fig:agot-steps}. Here, the "layer 0" is generated in step (i) and evaluated to completion in steps (ii) and (iii). The evaluation process is identical for every layer in AGoT, including any nested graphs, except special cases for the first and last layers, as in steps (i)-(iii) and (ix)-(x). 

Before articulating special cases, we consider the general one. In general, the evaluation of a single layer is complete when the evaluation of all its nodes is complete. All nodes start in the "new" state and progress through one of two trajectories:

\begin{enumerate}
    \item The node is classified as "complex" and its content directly informs a new, nested AGoT process.
    \item The node is \textit{not} classified as complex and gets evaluated directly.
\end{enumerate}

Nested AGoT processes stemming from case (1) above are still subject to the aforementioned evaluation process, with the content of the complex node informing the new nested graph in a manner analogous to how the initial query informs the first layer of the top-most graph. In all graphs, the next layer of nodes and edges is then added once the current one is completed.

Regarding special cases, the very first layer of a graph (or nested graph) is distinguished by the fact that its generation is informed by the initial query (or by the contents of the associated complex node). Initial layers are also generated without any explicit dependencies (\textit{i.e.} edges), although every node in the initial layer of a nested graph is implicitly a "child" node of the associated complex node. \autoref{fig:agot-hero} most clearly illustrates the global node and edge sets of a sample AGoT instance with depth 1 recursion.

\subsection{Dynamic frameworks}

The remainder of this work introduces a detailed mathematical formalism for AGoT, works through illustrative examples, presents benchmarking results across diverse tasks, and concludes with a discussion of future research directions.

Dynamism in the context of inference or generation frameworks refers to a framework's ability to adapt to individual tasks or datasets through automatic adjustments to its own structure, prompting strategies, or other aspects of the general methodology. Apart from CoT, all frameworks discussed thus far (and depicted in \autoref{fig:agot-hero}) are dynamic to varying degrees. Dynamic inference frameworks~\citep{prasad2024adaptasneededdecompositionplanning, ning2024dgotdynamicgraphthoughts, zhu2024redeltoolkitllmpoweredrecursive, radha2024iterationthoughtleveraginginner} represent some of the most performant approaches to date.

Designed to be adaptive and generalizable, dynamism in AGoT hinges on two principal degrees of freedom; (1) the ability to generate any number of new nodes (up to a user-specified limit) per layer, with arbitrary edges; and (2) the ability to recursively apply itself as necessary. In the process of generating nodes to decompose tasks and sub-tasks, AGoT also generates a unique \textit{strategy} to guide and focus each layer's objective. Additionally, AGoT is able to recognize high-quality responses and self-terminate, reducing unnecessary branching.

To validate AGoT as a general-purpose framework, we tested its performance on several difficult datasets belonging to "reasoning", "retrieval", and "explorative" categories. These results are summarized in \autoref{tab:performance_comparison} in \autoref{sec:results}.

AGoT is specifically designed to be both adaptive and generalizable. Its dynamism hinges on two key capabilities: the ability to generate an arbitrary number of new nodes per layer (subject to user-specified limits), and the capacity to recursively apply its reasoning process when necessary. During the generation of nodes for decomposing tasks and sub-tasks, AGoT concurrently formulates a unique strategy to guide each layer\'s objectives. Additionally, it can recognize high-quality responses and self-terminate, thereby reducing unnecessary branching. To validate AGoT as a general-purpose framework, we have tested its performance on several challenging datasets spanning reasoning, retrieval, and explorative problem-solving tasks, with results summarized in \cref{tab:performance_comparison} in \cref{sec:results}.

The remainder of this work introduces a mathematical formalism for AGoT, works through an example, discusses benchmarking results, and provides concluding remarks.

\begin{figure*}[ht]
\centering 
\begin{minipage}{0.95\textwidth}
    \begin{algorithm}[H]
\caption{Adaptive Graph of Thought $\mathbf{AGoT}(q,h,G_{h_p})$}
\label{alg:agot}
\renewcommand{\algorithmicrequire}{\textbf{Input:}}
\renewcommand{\algorithmicreturn}{\textbf{Output:}}
\begin{algorithmic}[1]
\Require Query $q \in \mathcal{Q}$, position index $h \in \mathcal{H}$, the parent graph $G_{h_p}$.
\State $V_h \gets \{\emptyset\}, E_h \gets \{\emptyset\}, F_h \gets \emptyset$
\State $G_h \gets (V_h, E_h, F_h)$ \Comment{initialize nested graph}
\State $d \gets |h|$ \Comment{get current depth}
\State $h' \gets (h, (0,0))$ \Comment{set starting heritage in $G_h$}
\Statex
\For{$l = 0, 1, \dots, l_\text{max}-1$} \Comment{loop layer by layer}
    \If{$l = 0$ \textbf{and} $d = 0$}
        \State $(\{t_{h'}\}, \sigma_0) \gets \mathbf{T}_{\emptyset}(q, n_\text{max})$ 
        \Comment{generate initial thoughts}
    \ElsIf{$l = 0$}
        \State $(\{t_{h'}\}, \sigma_0) \gets \mathbf{T}_0(q, G_{h_p}, n_\text{max})$ 
        \Comment{generate initial thoughts for nested graph}
    \Else
        \State $(\{t_{h'}\}, \sigma_l, E(V_h, \{v_{h'}\})) \gets \mathbf{T}(q, G_{h}, n_\text{max})$
        \Comment{generate thoughts, strategy, \& new edges}
        \State \textbf{append}($E_h$, $E(V_h, \{v_{h'}\}$) \Comment{update edge set}
    \EndIf
    \Statex
    \If{$t_{h'}$ is final}
        \State $F_h \gets \mathbf{Eval}(t_{h'}, G_h)$ 
        \Comment{evaluate the thought}
        \State \Return $F_h, G_h$ \Comment{\textbf{return final answer and graph}}
    \EndIf
    \Statex
    \For{$t_{h'} \in \{t_{h'}\}$}
        \If{$\mathbf{C}(t_{h'},G_h) = 1$ \textbf{and} $d < d_\text{max}$}
            \State $a_{h'}, G_{h'} \gets \mathbf{AGoT}(t_{h'}, h', G_h)$ 
            \Comment{recursive call}
        \Else
            \State $a_{h'} \gets \mathbf{Eval}(t_{h'}, G_h)$ 
            \Comment{evaluate the thought}
        \EndIf
    \EndFor
\EndFor
\Statex
\State $F_{h'} \gets \mathbf{Eval}(\mathbf{\Phi}(G_h), G_h)$ 
\Comment{final answer from final thought}
\State \Return $F_h, G_h$ \Comment{\textbf{return final answer and graph}}
\end{algorithmic}
\end{algorithm}

\end{minipage}   
\end{figure*}

\section{Mathematical formalism}
\label{sec:agot-formalism}




\subsection{Structure and indexing}\label{subsec:structure-indexing}
Every node in AGoT is uniquely defined by its heritage, a sequence of nodes that distinguishes a particular subgraph. Note that a heritage is \textit{not} necessarily the same as a node's "ancestry" in the usual DAG context because a heritage does not necessarily correspond to an edge-traversal path.

Letting $L$ be the set of layer indices and $N$ the set of node indices in an arbitrary layer, we define the set of \textit{local} indices at each nesting depth $d \leq d_\text{max}$ by $S = L \times N$. Every heritage then belongs to the following set
\begin{align}
    \mathcal{H} = \bigcup_{d = 0}^{d_\text{max}} S^d
\end{align}
where $S^d$ is the $d$-fold Cartesian product of $S$. Members of $\mathcal{H}$ are understood as sequences of $(d+1)$ node positions $s_i$
\begin{align}
    h &= (s_0, s_1, \dots, s_{d}) \\
    s_i &= (l_i, n_i)
\end{align}
where $0\leq d \leq d_\text{max}$. The indices $l_i \in \mathbb{N}$ and $n_i \in \mathbb{N}$ are the layer (\textit{i.e.} topological generation) and node label, respectively, which uniquely identify a node at nesting depth $i \leq d_\text{max}$. Any node in an entire AGoT instance can then be denoted by $v_h$ in reference to a heritage $h$ that traces a path to the node's position, $s_d$, via interfacial "complex" nodes $s_{i<d}$ that connect nested graphs across AGoT depths. In other words, the preceding sub-sequence $h^\prime = (s_0,\dots,s_{d-1})$ can be understood as the heritage of a graph or nested graph, wherein $s_d$ identifies a particular node.

The top-level AGoT graph is denoted by $G_{\emptyset}$ where $\emptyset$ represents the empty heritage. To proceed, we also define $\mathcal{G}$ as a class of hierarchical graphs; $\mathcal{Q}$ as the set of all possible queries; $\mathcal{A}$ as the set of all possible answers; $\Sigma$ as the set of all possible strategies; and $\mathcal{T}$ as the set of all possible tasks.

\begin{figure*}[h!]
    \centering
    \includegraphics[width=\linewidth]{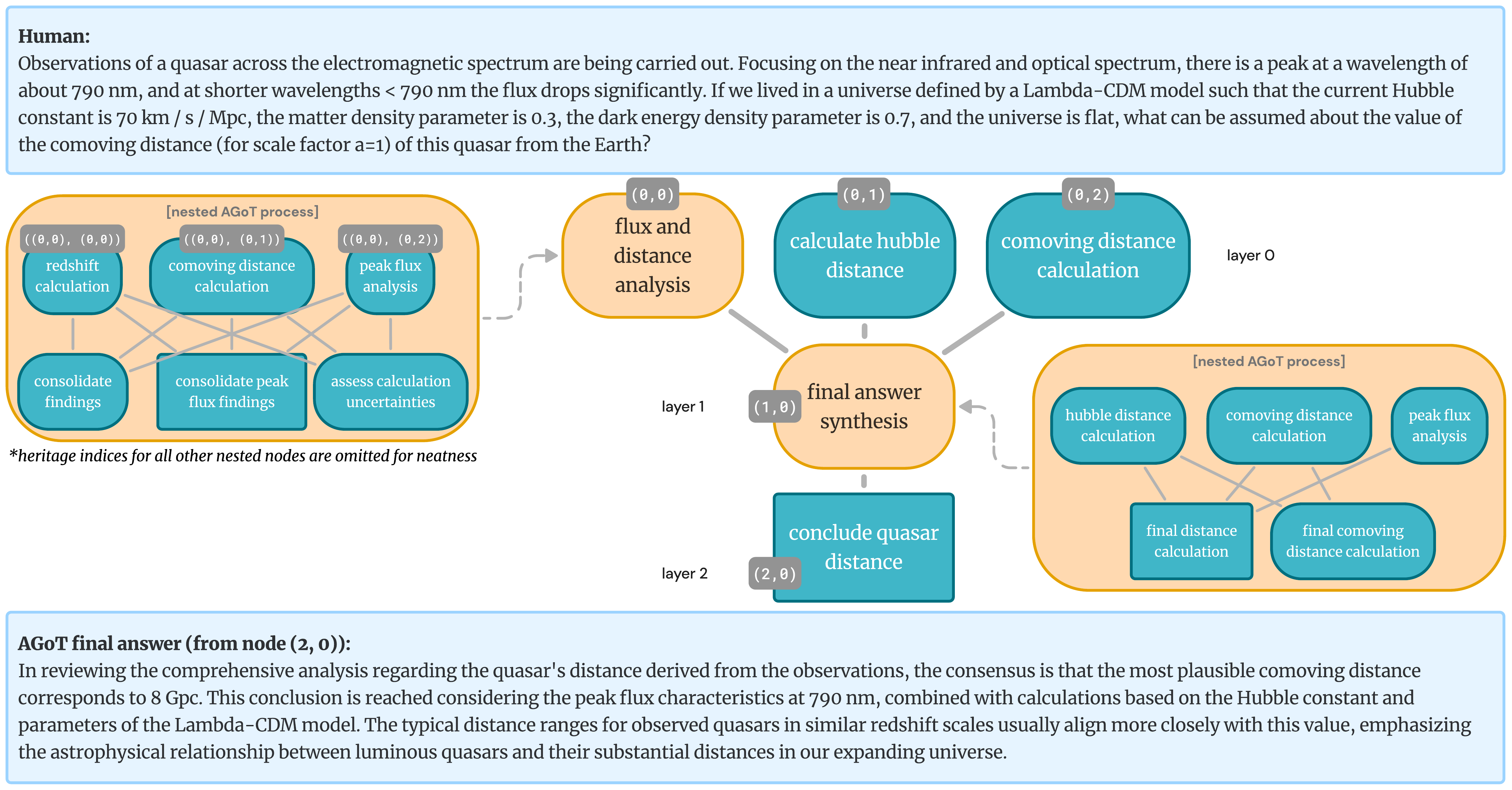}
    \caption{Diagram illustrating a final AGoT state after evaluation of a technical problem from GPQA. Grey labels on the top-level graph identify the single positions that comprise the heritages of these nodes. Each layer in the top graph is labeled from 0 to 2, while layer and heritage labels are omitted from nested graphs for neatness. Text inside each node indicates the generated thought's "title", but does \textit{not} uniquely identify any node's content.}
    \label{fig:agot_example}
\end{figure*}

We now associate with each node $v_h$ either the empty graph (denoted by $\emptyset$) or a nested graph,
\begin{align}
    G_h = (V_h, E_h, F_h)
\end{align}
defined by sets of nodes and edges, $V_h$ and $E_h$, as well as a final answer $F_h$, such that any node $v_h$ is exactly specified by
\begin{align}
    v_h = (t_h, \sigma_{l_i}, a_h, G_h)
\end{align}
where $t_h \in \mathcal{T}$ is the thought for the node, $\sigma_{l_i} \in \Sigma$ is the strategy for its layer, and $a_h \in \mathcal{A}$ is the answer or response to thought.

Since AGoT is structured recursively, by expanding complex nodes into nested graphs, the global node and edge sets are:
\begin{align}
    \mathcal{V} = \bigcup_{h \in \mathcal{H}}{v_h} && \mathcal{E} = \bigcup_{h \in \mathcal{H}} E_h && G = \bigcup_{h \in \mathcal{H}} G_h
\end{align}

We denote by $d_\text{max}$, $l_\text{max}$, and $n_\text{max}$ the arbitrary limits on maximum recursion depth, the number of layers, and the number of nodes per layer; noting that the \textit{minimum} depth can be 0 while layer and node counts must be at least 1.

\subsection{Actions on graph data}
Taking $\mathcal{Q}$ to be the set of all input queries and $\Theta$ to be the set of all $n$-tuples ($n = 1,2, \dots, N$) of thoughts,
\begin{align}
    \Theta = \mathcal{T} \cup \mathcal{T}^2 \cup \dots \cup \mathcal{T}^{|N|},
\end{align}

we describe crucial actions in AGoT with the mappings:

\begin{flalign}
    \mathbf{T}_{\emptyset}&: \mathcal{Q} \times N  \to \Theta \times \Sigma \label{f:T_empty}\\
    \mathbf{T}_0&: \mathcal{Q} \times N \times G \to \Theta \times \Sigma \\
    \mathbf{T}_e&: \mathcal{Q} \times N \times G \to \Theta \times \Sigma \times \mathcal{E} \\
    \mathbf{C}&: \mathcal{T} \times G \to \{0,1\} \label{f:C}\\
    \mathbf{Eval}&: \mathcal{T} \times G \to \mathcal{A} \\
    \mathbf{\Phi}&: G \to \mathcal{T} \label{f:Phi}
\end{flalign}

Among these, $\mathbf{T}_\emptyset$, $\mathbf{T}_0$, and $\mathbf{T}_e$ are the actions that map inputs to new thoughts. Specifically, $\mathbf{T}_\emptyset$ operates on initial queries to generate initial thoughts (e.g. \autoref{fig:agot-steps} (i)), while $\mathbf{T}_0$ operates on complex thoughts to generate initial thoughts in the first layer of nested graphs. $\mathbf{T}_e$ operates on all other thoughts not targeted by the prior two mappings. Note that the output space of $\mathbf{T}_e$ includes a new thought, a strategy, \textit{and} new edges, because $\mathbf{T}_e$ describes input-to-thought mapping in the most general case. The remaining mappings, $\mathbf{C}$, $\mathbf{Eval}$, and $\mathbf{\Phi}$ describe complexity checks, thought evaluation, and final thought generation respectively. With this definition on hand, AGoT is defined as pseudo-code in Algorithm \ref{alg:agot}.

\section{Example} \label{sec:example}
In this section, we provide a step-by-step description of an AGoT process that answers a sample GQPA~\citep{rein2023gpqa} question. Categorized as a "reasoning" task, this sample illustrates the decomposition of a difficult scientific question, as seen in \autoref{fig:agot_example} (next page).

Starting with the initial query (described as "human" input at the top of \autoref{fig:agot_example}), AGoT generates three key thoughts to comprise the initial layer ("layer 0"). Because these thoughts belong to the top graph, their heritage (see \autoref{subsec:structure-indexing}) contains only a single tuple, as shown in the small gray boxes adjacent to each top-level node. Upon evaluation of the initial layer, the node titled \textit{flux and distance analysis} is marked as complex, while the remaining nodes (index $(0,1)$ and $(0,2)$) are answered directly.\footnote{Wherever possible, AGoT actions happen asynchronously in our implementation.}

At this point, the nested AGoT process corresponding to node position $(0,0)$ must be completed before the next layer ("layer 1") is started. Having set $d_\text{max} = 1$ here, nested graphs themselves will not contain any complex nodes, so the first layer of thoughts --- whose heritages are $h_k = ((0,0), (0,k))$ for $k=0,1,2$ --- is generated then evaluated directly. Similarly, the second nested layer (indices omitted in \autoref{fig:agot_example}, $h_j = ((0, 0), (1, j))$ for $j=0,1,2$), which includes the final thought (\textit{consolidate peak flux findings}) for this nested graph. These thoughts represent a decomposition of the parent thought at $(0,0)$ and therefore address topics related to peak flux and cosmological distances. Upon completion of the first nested graph, the contents of its final answer inform the creation of the single node $(1,0)$, together with the answers of top-level nodes at $(0,1)$ and $(0,2)$. The new node at $(1,0)$ is then also identified as complex, initiating the second nested AGoT process shown in the bottom right of \autoref{fig:agot_example}. While the second nested graph contains some node titles also seen in the first, it is unlikely that these repeat the same information because the second nested graph is already informed by the answer from the first, via the edge from $(0,0)$ to $(1,0)$. Instead, the second nested graph is interpreted according to its parent complex node, as a subsequent iteration of the synthesis process to review and extract the key insights.

The final layer is "layer 2". Here, the thought contained in node $(2,0)$ is generated immediately after completion of the second nested graph. As the final node, its answer is the overall answer to the original input query. From \autoref{fig:agot_example} we see that $(2,0)$, informed by the entire graph preceding its own topological generation, has correctly determined that a comoving distance of 8 Gpc.

\section{Implementation and testing methodology} \label{sec:implementation}
Our implementation of AGoT utilizes an asynchronous LLM client and various instructions and response schemas to create the mappings (\ref{f:T_empty})-(\ref{f:Phi}). Among these combinations, each \texttt{(model,~instruction,~schema)} triplet equivalently comprises an "AI agent" with a specific (though not necessarily unique) interface and a directive to implement one of the mappings (\ref{f:T_empty})-(\ref{f:Phi}). With the mappings on hand, Algorithm \ref{alg:agot} provides an exact definition for AGoT.

We used the \texttt{gpt-4o-mini}~\citep{hurst2024gpt} model by OpenAI for all experiments in this work, including direct input-output (IO), CoT, AIoT, and AGoT. This model was accessed using an asynchronous chat client from the open-source \href{https://github.com/openai/openai-python}{\texttt{openai-python}} package, setting \texttt{temperature=0.3} for generation with AGoT and using the default settings otherwise. AGoT settings for $d_\text{max}$, $l_\text{max}$, and $n_\text{max}$ were set to 1, 3, and 3, respectively, for all experiments. While these values were found to produce a satisfactory balance of computation time to performance, we do not assert them as optimal here.

Before proceeding to our results, we emphasize that, unlike some existing inference frameworks~\citep{ning2024dgotdynamicgraphthoughts,zhu2024redeltoolkitllmpoweredrecursive,wang2024tdagmultiagentframeworkbased}, AGoT is neither tailored for any specific task nor pre-trained at all on any specific dataset.\footnote{An installable implementation of the AGoT framework and result data can be found at~\citep{multi_agent_llm_2024}.}

\section{Results} \label{sec:results}
This section details the outcomes of our benchmarking experiments. We tested AGoT on several datasets designed for reasoning, retrieval, and explorative abstract problem solving, finding better or significantly better performance on each, as compared to IO, CoT, and AIoT. Results on The Game of 24, which is a set of small, challenging problems on arithmetic operations, show AGoT reaching an average accuracy of 50\% on the 20 hardest puzzles, yielding an improvement of +400\% from over direct IO. This and the remaining results are discussed in the subsections that follow.

\subsection{Reasoning}
\subsubsection{GPQA} \label{subsubsec:gpqa}
GPQA Diamond~\citep{rein2023gpqa} consists of 198 highly-technical multiple-choice questions. Their systematic analysis reveals that conventional evaluation methods using fixed answer positions may not accurately reflect a model's true reasoning abilities, necessitating controlled experiments that account for positional bias~\citep{radha2025reasoning}. Following this insight, we conducted GPQA experiments both with and without shuffling the positions of multiple-choice answers. In the shuffled case, where positional cues are systematically controlled, AGoT with \texttt{gpt-4o-mini} proved +32.4\% more accurate than direct IO, achieving a best-overall total score of 49.5\% (98 correct out of 198 total answers). For consistent comparison with existing literature, which predominantly reports on unshuffled variants, we also evaluated on the standard dataset configuration where AGoT was +0.9\% more accurate than IO (see \autoref{tab:gpt-gpqa-results}), while still outperforming CoT and AIoT which yield negative results (see \autoref{tab:performance_comparison}). This marked performance differential between shuffled and unshuffled conditions aligns with ~\citet{radha2025reasoning}'s findings regarding position-dependent effects in reasoning tasks. The reduced gains on unshuffled data suggest potential saturation of position-specific patterns in the model's training, where self-reflexive analysis through frameworks like AGoT can provide only marginal benefits through more extensive exploration of this knowledge.

\subsubsection{GPQA comparisons} \label{subsubsec:gpqa-comparisons}
Noting empirically the importance of shuffled versus unshuffled answer positions, and for clarity in references to other works, we denote the default global "main set" of GPQA questions by GPQA$_\text{0}$; the GPQA Diamond subset by GPQA$_\text{D}$; and our Shuffled GPQA Diamond subset by GPQA$_\text{S}$.

\begin{table}[h!]
    \centering
    \begin{tabular}{l|lccr}
        \toprule
        \textbf{Dataset} & \textbf{Model} & \textbf{IO} & \textbf{AGoT} & $\mathbf{\Delta^\text{AGoT}_\text{IO}}$ \\
        \midrule
        GPQA$_\text{D}$ & \texttt{gpt-4o-mini} & 54.6 & \textbf{55.1} & +0.9\% \\
        \midrule
        \multirow{2}{*}{GPQA$_\text{S}$}
         & \texttt{gpt-4o-mini} & 37.4 & \textbf{49.5} & +32.4\% \\
         & \texttt{gpt-4o} & 39.4 & \textbf{57.6} & +46.2\% \\
        \bottomrule
    \end{tabular}
    \caption{GQPA answer accuracy of GPT models across dataset variations.}
    \label{tab:gpt-gpqa-results}
\end{table}

Compared to \texttt{gpt-4o-mini}, using AGoT with \texttt{gpt-4o} we found an even larger +46.2\% margin of improvement on GPQA$_\text{S}$. With this figure, AGoT approaches performance gains obtained via model "distillation" using DeepSeek-R1~\citep{deepseekai2025deepseekr1incentivizingreasoningcapability}, a computationally intense method based on fine-tuning and reinforcement learning. Taking \citet{dubey2024llama} (0-shot CoT, GPQA$_\text{0}$) and \citet{yang2024qwen2} (5-shot, GPQA$_\text{0}$) as a baseline, distillation improves reasoning with Llama3-70B and Qwen2.5-32B by +39.6\% and +29.4\%, respectively, while both model families improve around +46\% on average across all model sizes on GPQA$_\text{D}$. Reaching parity with model distillation constitutes an excellent outcome for reasoning with AGoT, especially considering that GPQA$_\text{S}$ is \textit{at least} as hard as GPQA$_\text{D}$ and that AGoT is a comparatively lightweight and modular "client-side" framework.


\begin{table*}[ht]
    \centering
    \begin{tabular}{l|lccrcr|cr}
        \toprule
        \textbf{Category} & \textbf{Dataset} & \textbf{IO} & \textbf{CoT} & $\mathbf{\Delta^\text{CoT}_\text{IO}}$ & \textbf{AIoT} & $\mathbf{\Delta^\text{AIoT}_\text{IO}}$ & \textbf{AGoT} & $\mathbf{\Delta^{\text{AGoT}}_\text{IO}}$ \\
        \midrule
        \multirow{2}{*}{\textbf{Reasoning}}
            & GPQA$_\text{D}$ & 54.6 & 49.0 &-10.3\% & 48.0 & -12.1\% & \textbf{55.1} & +0.9\% \\
            & GPQA$_\text{S}$ & 37.4 & 38.6 & +3.2\% & 39.4 & +5.4\% & \textbf{49.5} & +32.4\% \\
        \midrule
        \multirow{3}{*}{\textbf{Retrieval}} 
            & HotpotQA & 72.0 & 75.0 & +4.2\% & 76.0 & +5.6\% & \textbf{80.0} & +11.1\% \\
            & MoreHopQA & 55.0 & 63.0 & +14.5\% & 70.0 & +27.3\% & \textbf{72.0} & +30.9\% \\
            & HybridQA & 68.0 & 64.0 & -5.9\% & 77.0 & +13.2\% & \textbf{84.0} & +23.5\% \\
        \midrule
        \multirow{3}{*}{\textbf{Explorative}}
            & Crossword Letters & 18.6 & 19.2 & +3.2\% & 33.5 & +80.1\% & \textbf{34.7} & +86.6\% \\
            & Crossword Words & 2.5 & 3.0 & +20.0\% & 11.0 & +340.0\% & \textbf{11.1} & +344.0\% \\
            & Game of 24 & 10.0 & 20.0 & +100.0\% & 25.0 & +150.0\% & \textbf{50.0} & +400.0\% \\
        \bottomrule
    \end{tabular}
    \caption{Performance comparison between IO, CoT~\citep{wei2022chain}, AIoT~\citep{radha2024iterationthoughtleveraginginner}, and AGoT (this work) as inference frameworks with \texttt{gpt-4o-mini} across reasoning, retrieval, and explorative tasks. Values in brackets indicate percent improvement versus direct IO. AGoT attains the highest values across all rows. (GPQA$_\text{S}$ and GPQA$_\text{D}$ represent the shuffled and unshuffled versions of GPQA Diamond (\autoref{subsubsec:gpqa-comparisons}).}
    \label{tab:performance_comparison}
    
\end{table*}

\begin{table*}
    \centering
    \begin{tabular}{ll|ccc}
        \toprule
        \textbf{Dataset} & \textbf{Metric} & \textbf{IO} & \textbf{AIoT} & \textbf{AGoT} \\
        \midrule
         \multirow{3}{*}{HotpotQA} 
         & EM & \textbf{54} & 48 & 51 \\
         & F1 & 74.4 & 73.7 & \textbf{76.2} \\
         & LAAS & 72 & 76 & \textbf{80} \\
         \midrule
         \multirow{3}{*}{MoreHopQA}
         & EM & 41 & \textbf{52} & 49 \\
         & F1 & 52.9 & \textbf{63.6} & 60.0 \\
         & LAAS & 55 & 70 & \textbf{72} \\
         \midrule
         \multirow{3}{*}{HybridQA}
         & EM & \textbf{55} & 52 & 45 \\
         & F1 & 68.4 & \textbf{68.9} & 67.8 \\
         & LAAS & 68 & 77 & \textbf{84}\\
        \bottomrule
    \end{tabular}
    \caption{Performance comparison using IO, CoT~\citep{wei2022chain}, AIoT~\citep{radha2024iterationthoughtleveraginginner}, and AGoT (this work) as inference frameworks with \texttt{gpt-4o-mini} on three reasoning datasets across Exact Match (EM), F1, and LLM assisted accuracy scores (LAAS). The best result in each row is emphasized with bold characters.}
    \label{tab: Retrieval tasks}
\end{table*}

\subsection{Retrieval}
Our benchmarks in the retrieval category cover three multi-hop datasets, namely HotpotQA~\citep{yang2018hotpotqadatasetdiverseexplainable}, MoreHopQA~\citep{schnitzler2024morehopqamultihopreasoning}, and HybridQA~\citep{chen2021hybridqadatasetmultihopquestion}. In addition to exact match (EM) and F1 scores, we report for this category an LLM-assisted accuracy score (LAAS) obtained by requesting a binary response that indicates the semantic equivalence of two input strings. This metric is implemented under the reasonable assumption that \texttt{gpt-4o-mini} is capable of highly accurate text classification in this setting. For example, the strings \texttt{"12-16-1770"} and \texttt{"12-16-1707"} are almost identical but \textit{not} LAAS equivalent, whereas \texttt{"12-16-1770"} and \texttt{"16 December, 1770"} are in fact LAAS equivalent, given they represent the same date. For consistency and validation, the LAAS metric is reported for all inference frameworks alongside EM and F1 in \autoref{tab: Retrieval tasks}. We find that AGoT produces significantly better LAAS scores across all three retrieval datasets, while EM scores are most often highest for IO, and F1 scores are most often highest for AIoT~\citep{radha2024iterationthoughtleveraginginner}. The remainder of this subsection discusses results on each retrieval benchmark in greater detail.

\subsubsection{HotpotQA}
This dataset contains multi-hop questions based on Wikipedia content across various domains. Here, we report results on a random sample of 100 "hard multi-hop"~\citep{yang2018hotpotqadatasetdiverseexplainable} questions from HotPotQA. We note that previous work has achieved top accuracy of 81\%~\citep{gao2024metareasoninglargelanguage} over the entire dataset (\textit{i.e.} with nominally easier questions on average) whereas AGoT achieves 80\% accuracy on a subset of only nominally \textit{hard} problems. Looking at EM and F1 scores, direct reasoning and retrieval-augmented reasoning models have previously yielded top scores of $45\%$ and $57.3\%$~\citep{li2025searcho1agenticsearchenhancedlarge}, respectively, whereas AGoT with \texttt{gpt-4o-mini} achieves $51\%$ and $76.2\%$ on EM and F1 in this work.

\subsubsection{MorehopQA}
This dataset is similar to HotpotQA, except it emphasizes a more generative approach instead of pure retrieval. To avoid encouraging the test subject to pick correct answers from the context or to simply recombine contextual information into a correct answer, clues for each question in MorehopQA must be "understood" in order to synthesize relevant information that informs the correct answer. (See  ~\citet{schnitzler2024morehopqamultihopreasoning} for examples.) Based on results in \autoref{tab: Retrieval tasks}, MorehopQA is more challenging than a random sample of 100 nominally "hard" HotpotQA questions and corresponds to the lowest retrieval scores overall, for all inference frameworks considered.

In terms of LAAS, performance improves monotonically when comparing IO to CoT to AIoT to AGoT. However, while both AIoT and AGoT significantly out-perform the former two methods, the difference between them is marginal on MorehopQA, with AGoT performing slightly better at 72\% versus 70\% for AIoT. The LAAS score for AGoT on MorehopQA also represents the largest margin of improvement over direct IO across all three retrieval tasks (see \autoref{tab:performance_comparison}). Regarding EM and F1 scores, performance differences between AIoT and AGoT are again marginal, with AIoT performing slightly better.

\subsubsection{HybridQA}
This final multi-hop dataset~\citep{chen2021hybridqadatasetmultihopquestion} contains both structured and unstructured data from Wikipedia articles, including data tables as well as text passages. HybridQA, accordingly, places a stronger emphasis on the subject's ability to parse data presented in formats other than natural language. On this dataset, EM, F1, and LAAS scores were highest for IO, AIoT, and AGoT respectively, with CoT yielding a \textit{negative} improvement. The LAAS score for AGoT was its highest overall across all datasets in the retrieval category, and second-highest in terms of relative improvement versus direct IO.

\subsection{Explorative datasets}
We use the term "explorative" to denote tasks that are expected to benefit from combinatorial searches and explicit considerations of multiple alternatives. In this work, we consider a \href{https://en.wikipedia.org/wiki/24_(puzzle)}{Game of 24} dataset and a mini-crosswords dataset for testing AGoT in an explorative setting. Solutions here are represented by one of many combinations of arithmetic operations on a set of integers (Game of 24) or one of many combinations of intersecting alphabetic characters (Mini-crossword). Both datasets are considered challenging for LLMs because they subvert typical natural language processing goals due to their character-wise processing requirements. Despite this, we recognize good performance on such datasets to be essential for true general-purpose reasoning with LLMs. Results on Game of 24 and Mini-crosswords are discussed in the remainder of this subsection.

\subsubsection{Mini-crosswords}
This dataset contains small, generic crossword puzzles. Each puzzle includes a list of clues corresponding to intersecting horizontal and vertical words, as usual. Note that crossword solution is necessarily explorative due to dependencies between words. Individual word clues do not necessarily suggest unique solutions \textit{per se}, such that two or more possible answers may appear equally valid throughout the came. In other words, solution uniqueness is enforced holistically in crosswords, so (at least for human solvers) considering alternatives along the way is a necessity.

Following \citet{yao2024tree}, we consider both latter and word accuracy to grade performance on mini-crosswords. Letter accuracy is computed as the number of letters in the answer whose identity and position are in agreement with the known solution, divided by the total number of letter positions. We also report the \textit{word} accuracy as the number of words in the answer whose positions agree with the known solution. Note that a perfect word accuracy implies a perfect letter accuracy and \textit{vice versa}. However, letter accuracy is still a weaker measure in this setting, since high letter accuracy does not imply high (or even non-zero) word accuracy.

\begin{figure}[t]
    \centering
    \includegraphics[width=0.97\linewidth]{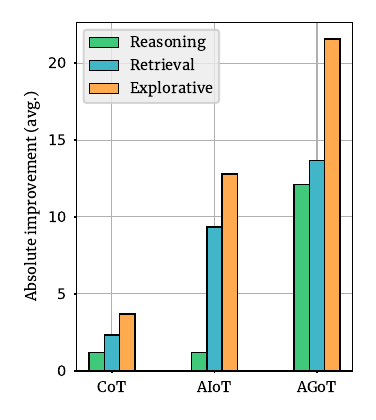}
    \caption{Average absolute difference in performance score for CoT, AIoT, and AGoT versus IO, using \texttt{gpt-4o-mini}. Reasoning category excludes the unshuffled GPQA$_\text{D}$ results (see \autoref{subsubsec:gpqa}).}
    \label{fig:performance comparison}
\end{figure}

Our results in \autoref{tab:performance_comparison} indicate that both AIoT and AGoT lead to much better performance on mini-crosswords, with over +80\% and +340\% improvements in letter and word accuracy, respectively, compared to direct IO. Notably, the improvement margin is much larger for AGoT. Improvements were also significant for CoT over direct IO, though both AIoT and AGoT score significantly better still. In absolute terms, however, solution accuracy is low across the board. Overall, AGoT achieved the highest letter and word accuracies of 34.7\% and 11.1\%, respectively.

Recall that AIoT~\citep{radha2024iterationthoughtleveraginginner} is a dynamic inference framework based on iterative dialogue and self-reflexive guidance. AGoT, while also dynamic, is more structurally sophisticated and designed specifically for decomposition. Based on their near-identical performance on mini-crosswords, we are unable to infer that AGoT is especially adaptable to solving mini-crosswords. The possibility therefore remains that \textit{any} sufficiently persistent framework can yield a relatively large baseline improvement for this task.

\subsubsection{Game of 24}
A single instance of Game of 24 involves finding a combination of unique arithmetic operations that, when applied to a specific set of 4 integers, produces an expression equaling exactly 24. Scoring on this dataset in binary, such that every answer is either entirely correct or entirely incorrect. We tested AGoT and other inference frameworks on the 20 most difficult questions in the dataset (with difficulty measured by the known success rate for human solvers). At the time of writing, we are aware of at least one preceding result~\citep{li2025searcho1agenticsearchenhancedlarge} on Game of 24, wherein ToT with \texttt{gpt-4} achieved 41\% accuracy on a random sample of 100 questions. In comparison, our experiments yield an accuracy of 50\% on the 20 hardest questions using AGoT with \texttt{gpt-4o-mini}. This represents the best result by far among the inference frameworks considered in \autoref{tab:performance_comparison}, suggesting that AGoT's recursive structure may be especially useful for mathematical problem solving.

\section{Discussion}
Our results indicate that AGoT is significantly more accurate than IO and CoT, and as or more accurate than AIoT across every task considered. AGoT dominates most strongly among tasks in the reasoning category, followed by the retrieval category (where LAAS is employed as the accuracy measure). An interesting caveat is the apparent saturation of accuracy values across frameworks for \textit{unshuffled} GPQA Diamond (see \autoref{tab:performance_comparison}), for which the first multiple-choice answer is always correct by default. Among the tests conducted, the unshuffled GPQA$_\text{D}$ was the only case where direct IO was the runner-up to AGoT, outperforming both CoT and AIoT. Comparing direct IO results for the shuffled and unshuffled cases then suggests a strong in-built bias for answering GPQA Diamond questions with the first multiple-choice option. However, due to the proprietary nature of the model, we can not verify conclusively whether \texttt{gpt-4o-mini} or \texttt{gpt-4o} was exposed to GPQA Diamond questions (in their default, unshuffled state) during training.

AGoT produces much better results than AIoT in the Game of 24, despite doing only marginally better on the mini-crossword task. In both these tasks, correctness hinges on the position of individual characters in the final response. However, Game of 24 concerns exclusively symbols and numeric characters, whereas Mini-Crosswords concerns alphabetic letters. We therefore speculate that the extensive natural language training of LLMs (together with established approaches for text tokenization) can introduce a detrimental bias with respect tasks like mini-crosswords solution that feature non-horizontal arrangements of common words. On the other hand, this bias would be absent for mathematical puzzles like Game of 24, which do not involve any text re-orientation.

\begin{figure}[t]
    \centering
    \includegraphics[width=0.97\linewidth]{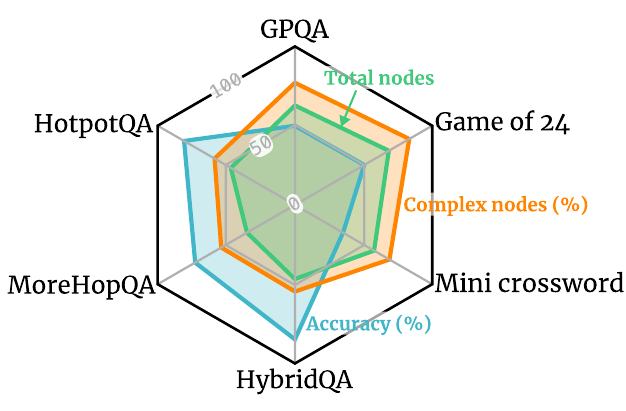}
    \caption{Total nodes and the percentage of complex nodes in AGoT with \texttt{gpt-4o-mini} for each task.}
    \label{fig:task-performance-breakdown}
\end{figure}

Because recursion in AGoT is triggered by complexity checks (as per \autoref{f:C} and Algorithm \ref{alg:agot}), the proportion of "complex nodes" in an solution serves as a proxy for the perceived conceptual difficulty of a given task. \autoref{fig:task-performance-breakdown} indicates that Game of 24, followed by GPQA and Mini-Crosswords, produced the three highest proportions of complex nodes, thus distinguishing non-retrieval tasks as those warranting the greatest degree of decomposition. The trend in total nodes (complex or otherwise) is the same, since the presence of complex nodes (due to spawning nested graphs) encourages a greater number of nodes overall.

In terms of average accuracies across the task categories, each framework is most accurate in the retrieval category, followed by reasoning, and finally explorative tasks. In terms of average relative improvement over IO, \autoref{fig:performance comparison} clearly highlights AGoT as having the greatest positive effect. Due largely to Game of 24, the average score in the explorative category represents the largest individual relative improvement for AGoT. We infer from these results that the AGoT, as a general-purpose framework, significantly improves the reasoning, retrieval, and explorative problem solving capabilities of the underlying LLM.

\subsection{Directions for future work}
It is not our intention to assert that our specific implementation of AGoT, based on the core definition in \autoref{sec:agot-formalism}, is by any means optimal. That is to say, any implementation of Algorithm \ref{alg:agot} together with an LLM-based implementation of Eqs. (\ref{f:T_empty})-(\ref{f:Phi}), whether in an agentic manner or otherwise, may very well produce a better version of AGoT with respect to the benchmarks considered. There is a large set of AGoT-compatible response schemas and instruction sets, so a more effective implementation \textit{vis a vis} the reported results in Tables \ref{tab:performance_comparison} and \ref{tab: Retrieval tasks} is not unlikely. For example, better choices for the arbitrary $d_\text{max}$, $l_\text{max}$, and $n_\text{max}$ parameters (here set to 1, 3, and 3, respectively) may yet be discovered through various optimization techniques. 

From the agentic perspective, recent studies~\citep{li2024more} have suggested that abstractions using larger pools of AI agents lead to better reasoning performance, up to a saturation limit. As discussed in \autoref{sec:implementation}, AGoT can be understood as being driven by the actions of six unique agents defined by Eqs. (\ref{f:T_empty})-(\ref{f:Phi}). An immediate avenue to investigate the potential benefits of additional agents exists in the thought mapping functions $\mathbf{T}_\emptyset$, $\mathbf{T}_0$, and $\mathbf{T}_e$, where both strategy and edge generation are suitable for re-delegation to specialized agents. Flexibility in the agentic perspective permits the use of LLMs other than \texttt{gpt-4o-mini}, as required or necessary for effective specialization. Beyond this, fine-tuning or reinforcement learning~\citep{wang2024reinforcementlearningenhancedllms} to beneficially bias (for example) edge selection or complexity checks may improve both AGoT performance and efficiency.

AGoT will also benefit form the incorporation of techniques to reduce LLM hallucination~\citep{tonmoy2024comprehensivesurveyhallucinationmitigation}. As with all iterative frameworks, AGoT readily enters situations where it re-digests its own responses. In the worst case scenario, this may lead to positive feedback loops that amplify incorrect or even nonsense responses; although even non-catastrophic hallucination wastes resources and may impact performance.

\section{Conclusion}

In this work, we introduced an Adaptive Graph of Thoughts (AGoT), demonstrating that careful structuring of the inference process can match the benefits of computationally intensive RL-based reasoning training methods. Using a variety of benchmark datasets, we demonstrated AGoT's superior performance in reasoning, retrieval, and explorative problem solving (see Tables \ref{tab:performance_comparison} and \ref{tab: Retrieval tasks} for summary of results) in comparison to existing methods. In contrast with some recent dynamic inference frameworks~\citep{ning2024dgotdynamicgraphthoughts,zhu2024redeltoolkitllmpoweredrecursive,wang2024tdagmultiagentframeworkbased}, these results are achieved in AGoT without pre-training, algorithmic tailoring, nor hyper-parameter optimization for any specific goal. In particular, AGoT improves reasoning with \texttt{gpt-4o} and \texttt{gpt-4o-mini} by relative margins comparable to base model distillation with DeepSeek-R1~\citep{deepseekai2025deepseekr1incentivizingreasoningcapability}.

Compared to our previous work on AIoT~\cite{radha2024iterationthoughtleveraginginner}, results on GPQA~\cite{rein2023gpqa} indicate that AGoT exhibits significantly better reasoning in comparison. Similarly, results on Game of 24 indicate that AGoT is approximately \textit{twice} as effective as AIoT at mathematical puzzle solving. AGoT's performance on mini-crosswords, however, showed no significant improvement, which suggests a potential blind spot in the approach of both methods. Despite the large relative improvement, datasets in the explorative category yielded the lowest \textit{absolute} scores (see \autoref{fig:performance comparison}), remaining a challenging category of tasks for LLMs (see \autoref{fig:task-performance-breakdown}).

As a framework for high-level LLM interaction~\citep{openaiblackbox,zhuang2024hydramodelfactorizationframework,pmlr-v235-sun24p} AGoT's performance benchmarks correlate strongly with the capabilities of the underlying model(s) that execute its crucial actions (\ref{f:T_empty})-(\ref{f:Phi}). With OpenAI's \texttt{gpt-4o-mini}~\citep{openai2024gpt4ocard}, we found that AGoT provides,  relative to direct inference (IO), an approximate +30\% performance boost in reasoning, +22\% in retrieval, and a whopping +277\% for explorative problem solving, on average. With the larger \texttt{gpt-4o}, reasoning performance improved even more, approximately +46\%.

The key advantage of AGoT lies in its ability to achieve these improvements through dynamic decomposition and structured recursion, rather than through computationally expensive model modifications. This approach provides a more accessible path to enhanced LLM performance, particularly valuable in resource-constrained settings or when model fine-tuning is impractical. As LLMs continue to evolve, frameworks like AGoT that can effectively augment model capabilities without requiring architectural changes or additional training will become increasingly important for practical applications.

\bibliographystyle{plainnat}
\bibliography{bib}

\end{document}